\documentclass[letterpaper, 10 pt, conference]{ieeeconf}  

\IEEEoverridecommandlockouts                              

\usepackage{blindtext}
\usepackage{siunitx}
\usepackage{graphicx}
\usepackage[belowskip=-5pt,aboveskip=5pt]{caption}
\usepackage{hyperref}
\usepackage{multirow}

\usepackage{tikz}

\def\checkmark{\tikz\fill[scale=0.4](0,.25) -- (.2,0) -- (0.9,.65) -- (.2,.15) -- cycle;} 

\title{\LARGE \bf
Improving 3D Object Detection for Pedestrians with Virtual Multi-View Synthesis Orientation Estimation
}

\author{Jason Ku, Alex D. Pon, Sean Walsh, and Steven L. Waslander
\thanks{Jason Ku, Alex D. Pon, Sean Walsh, and Steven L. Waslander are with the Institute for Aerospace Studies, University of Toronto, 4925 Dufferin St, North York, ON, Canada.
{\tt\small \newline 
kujason.ku@mail.utoronto.ca, alex.pon@mail.utoronto.ca, \newline
sean.walsh@mail.utoronto.ca, stevenw@utias.utoronto.ca}}%
}

\begin{document}

\maketitle
\thispagestyle{empty}
\pagestyle{empty}

\begin{abstract}
Accurately estimating the orientation of pedestrians is an important and challenging task for autonomous driving because this information is essential for tracking and predicting pedestrian behavior. This paper presents a flexible Virtual Multi-View Synthesis module that can be adopted into 3D object detection methods to improve orientation estimation. The module uses a multi-step process to acquire the fine-grained semantic information required for accurate orientation estimation. First, the scene's point cloud is densified using a structure preserving depth completion algorithm and each point is colorized using its corresponding RGB pixel. Next, virtual cameras are placed around each object in the densified point cloud to generate novel viewpoints, which preserve the object's appearance. We show that this module greatly improves the orientation estimation on the challenging pedestrian class on the KITTI benchmark. When used with the open-source 3D detector AVOD-FPN, we outperform all other published methods on the pedestrian Orientation, 3D, and Bird's Eye View benchmarks.
\end{abstract}

\section{Introduction}
Deep neural networks have made remarkable progress on the task of 3D object detection such that they are robust enough to be deployed on autonomous vehicles. The KITTI~\cite{geiger_kitti} benchmark showcases the success of 3D object detection methods, especially on the car and cyclist classes, but also highlights areas needing improvement. The benchmark shows that existing 3D detection methods~\cite{ku_avod, yan2018second} are able to very accurately estimate the orientation of cars and cyclists, with Average Angular Errors (AAE)~\cite{kundu_3drcnn} less than \ang{7} and \ang{20}, respectively, whereas the average error for pedestrians is almost \ang{56}.

In this work, we address the task of 3D pose estimation for pedestrians with a focus on orientation estimation. This task is especially important for autonomous driving as this information is useful for tracking and is vital for predicting pedestrian behavior. Furthermore, it is important to incorporate orientation estimation into the detection pipeline, and not rely on a tracking method for this estimation, as the orientation of pedestrians waiting to cross the street must be inferred without the aid of motion cues.

\begin{figure}[t]
    \centering
    \includegraphics[width=0.9\columnwidth]{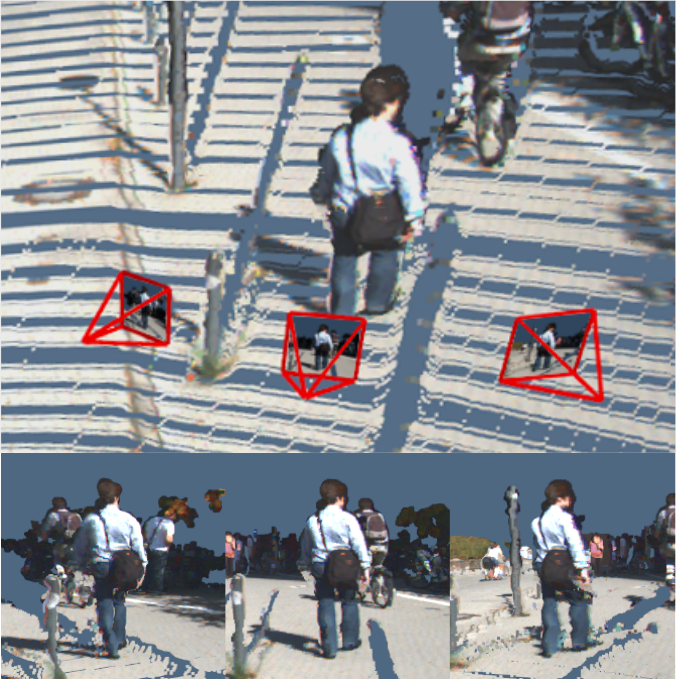}
    \caption{\textbf{Virtual Multi-View Synthesis.} The core idea of the method is to generate a set of virtual views for each detected pedestrian, and exploit these views in both the training and inference procedures to produce an accurate orientation estimation.}
  \label{fig_cover_photo}
\end{figure}

Image based detection methods have access to rich semantic information from RGB data. To accurately estimate orientation, these methods must extract fine-grained details of objects. However, extracting semantic information is challenging because of the varying scale and appearance of objects caused by the perspective transformation of the 3D scene into an image. Some methods~\cite{xiang_subcnn, tulsiani2015viewpoints} attempt to resolve this issue by using images at multiple scales to extract features. However, object appearance inconsistencies still exist from ROI cropping as noted in \cite{kundu_3drcnn}. In contrast, we propose to learn fine-grained information by rendering multiple viewpoints of objects through the placement of virtual cameras within the 3D scene at consistent locations relative to each object. Using these generated viewpoints retains a more consistent object appearance, as shown in Fig~\ref{fig_cover_photo}, which facilitates the learning of object orientation in our neural network.

\begin{figure}[t]
    \centering
    \includegraphics[width=1.0\columnwidth]{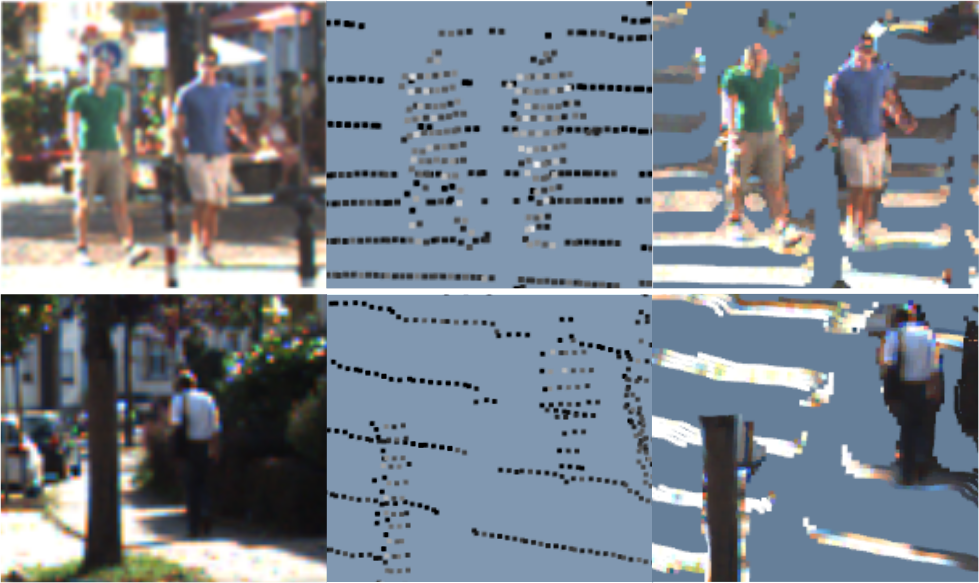}
    \caption{\textbf{Pedestrian Appearances at 20~m (top row) and 30~m (bottom row).} From left to right: RGB image, LiDAR scan colored by intensity, depth completed point cloud colored with corresponding RGB pixels. Even for humans, the classification of objects such as the tree, and the orientations of the pedestrians are not readily apparent in the LiDAR scan. In our method, rich semantic image features are preserved and fused directly with the point cloud, making it much easier to discern this information.}
  \label{fig_lidar_vs_depth}
\end{figure}

LiDAR methods can take advantage of accurate depth information to achieve robust localization. For the car and cyclist classes, the ratio of the 3D object length and width can be leveraged to simplify the orientation estimation problem~\cite{ku_avod}. However, LiDAR based methods also face challenges in extracting fine-grained semantic information of objects. The sparsity of LiDAR data limits the operational range of these methods, especially for smaller objects such as pedestrians. At longer distances, fine-grained information is lost as the sparsity of the LiDAR data becomes so severe that it can become difficult to distinguish between trees, poles, and pedestrians, leading to false positives. Fig.~\ref{fig_lidar_vs_depth} shows that even at shorter distances of 20 and 30 meters, with a high density 64 beam HDL-64E LiDAR, it is difficult for humans to discern meaningful orientation information due to the sparsity of the data. To solve this sparsity issue, we leverage the task of depth completion to generate dense point clouds, which allows for a one-to-one pixel-to-point incorporation of RGB image data. Also, inspired by F-PointNet's~\cite{qi_fpointnet} use of a 2D detector for accurate classification, we leverage a 2D detector for false positive suppression.

Moreover, orientation estimation performance is dependent on the amount of available training data. The training set must well represent the entire range of possible orientations, but labelled 3D data is expensive and time consuming to acquire~\cite{lee2018leveraging}. The KITTI~\cite{geiger_kitti} dataset has only 4500 pedestrian training instances with 3D labels, making it difficult to train neural networks that normally require much larger amounts of data to achieve good performance. A common solution for the scarcity of pose data is to leverage CAD models and additional annotations for more training data \cite{su2015render, chabot_deepmanta}. We, however, do not require the use of additional data sources or labels. To mitigate the low amount of data, we develop a virtual multi-view rendering pipeline to generate novel realistic data from the image and LiDAR inputs. This generated data is incorporated into our network during both training and inference. Detected objects are re-rendered from a set of canonical camera views so that objects maintain a more consistent appearance as compared to using the 2D crops obtained by the common Region of Interest (ROI) crop-and-resize procedure. At inference time, these additional viewpoints are used to determine a more accurate orientation estimate.

To summarize, our contributions in this work are as follows:
\begin{itemize}
    \item We propose a flexible module that can be adopted into 3D object detection pipelines to improve orientation estimation.
    \item We solve the issues of limited pose data and the challenges of obtaining fine-grain details from images and LiDAR data in a novel pipeline that uses virtual cameras to generate viewpoints on a colorized depth completed LiDAR point cloud.
    \item At the time of submission, when incorporated into the detection pipeline of the open-source AVOD-FPN~\cite{ku_avod} 3D object detector, our method ranks first compared to all other published methods on the KITTI~\cite{geiger_kitti} pedestrian Orientation, 3D, and Bird's Eye View benchmarks.
    \item Compared to other detection methods with comparable orientation estimation performance, our method runs over 8 times faster
\end{itemize}

\section{Related Work}
Orientation (yaw, pitch, roll) represents one of the components defining the 3D pose of an object. The introduction of 3D pose datasets~\cite{geiger_kitti, xiang2016objectnet3d, xiang2014beyondpascal, matzen2013nyc3dcars} has spurred a myriad of 3D pose estimation methods, and as a result many techniques for orientation estimation have been explored.

\textbf{Feature Extraction at Multiple Scales.} Previous works have recognized that accurate orientation estimation requires a feature extraction process that captures fine-grained semantic information of objects. Zhang et al.~\cite{zhang2016faster} identify that standard Faster R-CNN feature maps are too low resolution for pedestrians and instead use atrous convolutions~\cite{chen2018deeplab} and pool features from shallower layers. Pyramid structures including image pyramids~\cite{adelson1984imagepyramid} and feature pyramids~\cite{lin2017feature} have also been leveraged to obtain information from multiple scales. SubCNN~\cite{xiang_subcnn} use image pyramids to handle scale changes of objects, and \cite{ku_avod} highlight the importance of a pyramid structure for small classes such as pedestrians. Moreover, \cite{he2017mask_rcnn, ren_fasterrcnn, jiang2018acquisition} have shown the importance in how methods crop ROI features. Kundu et al.~\cite{kundu_3drcnn} note that the standard ROI crop can warp the appearance of shape and pose, and propose the use of a virtual ROI camera to address this. We instead obtain fine-grained details by using multiple virtual ROI cameras placed canonically around each object's detected centroid. As compared to \cite{kundu_3drcnn}, we not only use 2D RGB data, but also use a 3D point cloud to produce realistic novel viewpoints that maintain consistent object appearance.

\begin{figure*}[t!]
    \centering
    \includegraphics[width=1.0\linewidth]{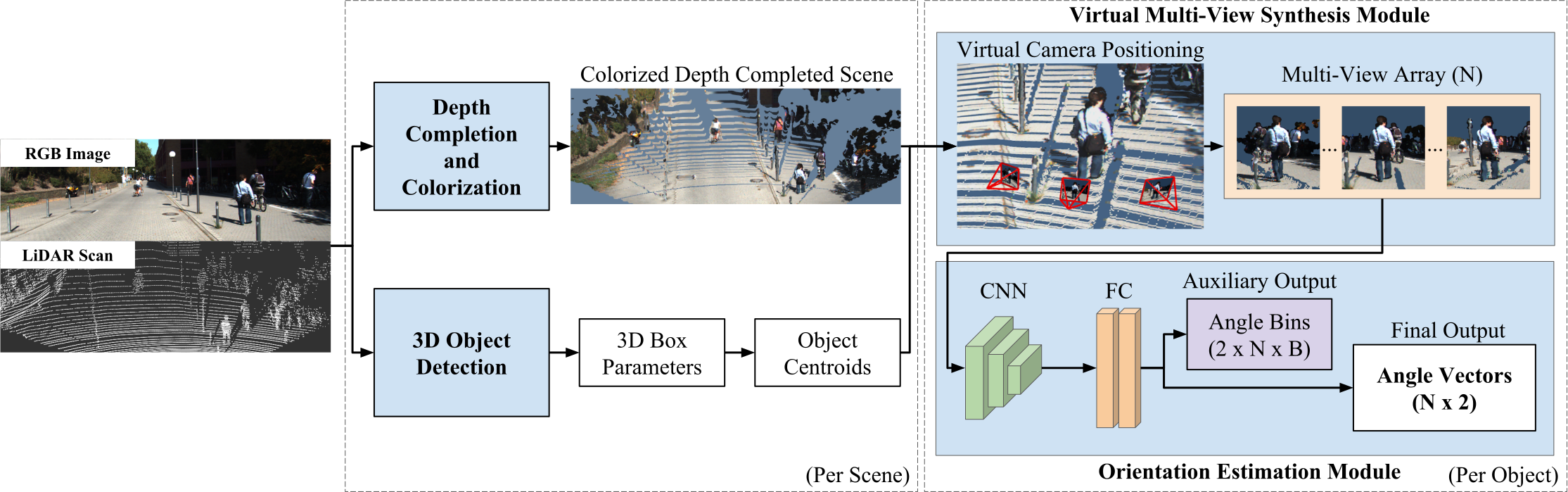}
    \caption{\textbf{Architecture Diagram.} A 3D detector is used to generate 3D detections, which are passed into the Virtual Multi-View Synthesis Module. The module places virtual cameras within the scene represented by a colorized depth completed LiDAR scan to generate $N$ novel viewpoints. Finally, the Orientation Estimation Module predicts the object orientation from the generated views.}
  \label{fig_architecture}
\end{figure*}

\textbf{Keypoints and CAD Models.} Using keypoint detections~\cite{toshev2014deeppose, tompson2015efficient, carreira2016human, shih2015part} and CAD models~\cite{pavlakos20176, wu2016single} have been shown to be effective in gaining semantic understanding of objects of interest. The use of 2D keypoint detections to estimate pose has been well studied as the Perspective-n-Point~(PnP) problem with many proposed solutions~\cite{lepetit2009epnp, lu2000fast, ansar2003linear}. More recently, \cite{pavlakos20176, wu2016single} use 3D CAD models and convolutional neural networks~(CNNs) to detect keypoints to learn 3D pose. Within the autonomous driving context, Deep MANTA~\cite{chabot_deepmanta} predicts vehicle part coordinates and uses a vehicle CAD model dataset to estimate 3D poses. CAD models have also been used to create additional ground truth labels. Su et al.~\cite{su2015render} argue the scarcity of training data with viewpoint annotations hinders viewpoint estimation performance, so they use 3D models to generate accurate ground truth data. Unlike the above methods, we do not require additional keypoint labels or external CAD datasets. We propose a general pipeline that leverages available data, and with our virtual cameras we generate novel high resolution viewpoints.

\textbf{Multi-view Learning.} Using multiple views has previously been shown to be effective in allowing neural networks to learn shape and pose information. Su et al.~\cite{su2015multi} render multiple views around an object from a CAD dataset, then predict shape based on each view's features. Others~\cite{tulsiani2017multi, tulsiani2018multi, mandikal2018capnet} use multiple views to ensure a projection consistency to learn shape and pose information. These methods tend to use CAD models, which segment the object of interest from the background, contain full $\ang{360}$ shape information, and allows perfect generation of data from any viewpoint. However, our application in autonomous driving only has access to LiDAR scans, which only provides input data from a single direction. We show that we are still able to exploit this data by carefully placing the virtual cameras within a certain operating region to maintain realistically rendered images, as shown in Fig.~\ref{fig_cam_placement}.

\textbf{Orientation Representation.} Most similar to our work are 3D pose estimation methods designed for autonomous driving scenarios. These methods have mainly focused on the representation of orientation and designing new loss functions. Pose-RCNN~\cite{braun2016posercnn} uses a Biternion representation for orientation as recommended by \cite{beyer2015biternion}. The monocular 3D object detection method Deep3DBox~\cite{mousavian_deep3dbox} proposes a \textit{angle bin} formulation that frames orientation estimation as a hybrid classification-regression problem. Here, orientation is discretized into several bins, and the network is tasked to classify the correct bin and to predict a regression offset. This formulation has been adopted by LiDAR methods including \cite{qi_fpointnet}. \cite{ku_avod} identify an ambiguity issue where identical 3D boxes are created despite having orientation estimates differing by $\pm \pi$ radians. They solve this issue by parameterizing orientation as an angle vector, while Yan et al.~\cite{yan2018second} approach this same issue with a sine error loss. We show in our ablation studies (Sec.~\ref{sec_ablation_angle_reg}) that parameterizing orientation as an angle vector while using the discrete-continuous angle bin formulation as an auxiliary loss is most effective.

\section{Pose Estimation Framework}
Fig.~\ref{fig_architecture} provides an overview of our pipeline for 3D pedestrian pose estimation. Given a set of 3D detections, each parameterized by a centroid $T=(t_x, t_y, t_z)$, dimensions $D=(d_x, d_y, d_z)$, and an Euler angle based orientation $O=(\theta, \phi, \psi)$, the objective is to provide a more accurate estimate of the orientation of the object of interest. For the KITTI benchmark, the pitch and roll are assumed to be zero, while only the yaw, or heading angle, $\theta_{yaw}$ is considered.

The core idea of our method is to generate realistic novel viewpoints of objects from a colored densified point cloud, and to use these viewpoints to extract rich semantic information of objects and thus perform more accurate orientation estimation. To start, we build upon the performance of existing 3D detectors by using 3D detections as centroid proposals. These proposals are processed by the Virtual Multi-View Synthesis module, which renders a set of novel views using a dense point cloud reconstruction. Importantly, these views are created in a set of canonical camera viewpoints, by placing virtual cameras at consistent locations relative to each object's centroid. The object views generated from these virtual cameras better preserves the 3D shape and appearance of the object as compared to taking ROI crops.

For each of the generated novel views, orientation is estimated by passing the views through a CNN followed by an orientation regression output head. The final orientation output is produced by merging the orientation estimates. Lastly, since the 3D detector used in the pipeline is chosen on a basis of having high recall, we use a robust 2D detector for false positive suppression.

\subsection{Virtual Multi-View Synthesis}

\textbf{Densified RGB Point Cloud Generation.} We first note that a LiDAR point cloud is simply a sparse representation of the underlying scene, and can be viewed omnidirectionally, with each view providing a unique visual representation of the scene. However the sparsity of LiDAR data only provides a low resolution perspective of the scene. We therefore take the portion of the LiDAR scan corresponding to the visible portion of the scene in the RGB image, and process it through a structure preserving depth completion algorithm~\cite{ku_ipbasic}. In particular, the LiDAR points are projected into the image using the provided $3 \times 4$ camera projection matrix $P_{cam}$ creating a sparse depth map $D_s$. The depth completion algorithm produces a dense depth map $D_d$, with each pixel representing a depth, that is then re-projected as a 3D point cloud. As the resulting point cloud comes from the dense depth map which has the same resolution as the RGB image, we next infuse the semantic information from the image by coloring each 3D point with its corresponding pixel from the RGB image. This colorized, dense reconstruction of the scene's point cloud resolves the low resolution of the LiDAR scan, and allows for more realistic novel views to be generated that preserve fine-grained semantic information, as shown in Fig.~\ref{fig_lidar_vs_depth}.

\textbf{Multi-View Generation.} Our goal is to learn fine-grained details of each object. The most straightforward process is to simply take ROI crops of the areas of the image corresponding to each object, but as noted by \cite{kundu_3drcnn} this common cropping procedure can lead to wildly different appearances of the same object. The use of object-centric coordinate frames has recently been shown to be effective for facilitating learning tasks~\cite{qi_fpointnet, rematas_soccer}, and as such, we design our module to use canonical viewpoints for estimation.

\begin{figure}[t]
    \centering
    \includegraphics[width=0.9\columnwidth]{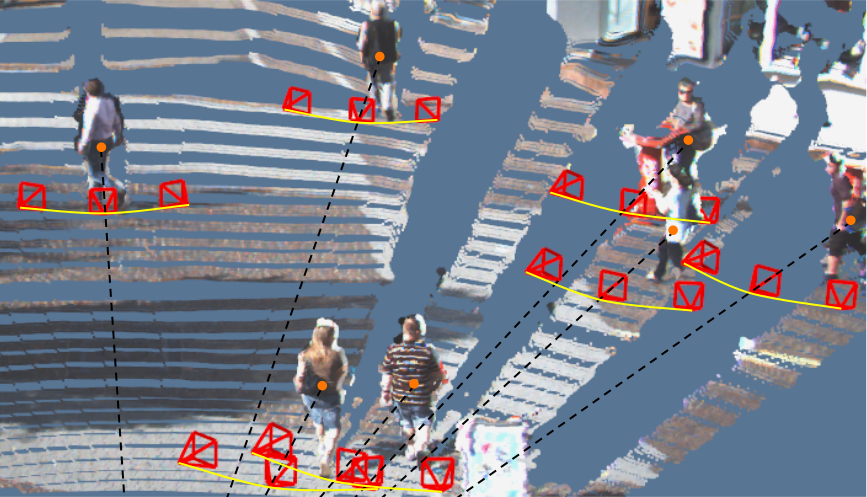}
    \caption{\textbf{Virtual Camera Placement.} Virtual cameras are placed at positions equidistant from each object centroid, with viewpoints ranging from \ang{-25} to \ang{25} relative to the ray from the original camera center to the object centroid, shown by the dotted black line. Here, only three of the eleven camera positions are shown.}
  \label{fig_cam_placement}
\end{figure}

To create our canonical camera viewpoints, we place virtual cameras in positions equidistant from the object centroid along $N$ equally spaced angle divisions. The camera locations are placed level with the center of the object of interest and between a range of angles, $\rho \in [-\rho_{max}, \rho_{max}]$ to the left and right of the horizontal viewing angle $\alpha$, defined by the ray from the original camera center to the object centroid, as shown in Fig.~\ref{fig_cam_placement}. The minimum and maximum angle of the viewpoints are chosen such that the generated views do not expose parts of the object that are not visible from the original camera viewpoint, which would make the object look unrealistic. $N$ views of the object are generated in a set of canonical viewpoints, which are evenly spaced along an arc of radius $r$ around the object. Each viewpoint generates an $H \times W$ ROI image that preserves appearance, as shown in Fig.~\ref{fig_cover_photo}. These rendered outputs densely fuse the point cloud and image information into a 3 channel RGB format, which allows the use of mature CNN architectures to be used for orientation estimation.

\subsection{Orientation Estimation}
The rendered ROI images are passed through a CNN to produce the final orientation estimations. Several methods \cite{mousavian_deep3dbox, qi_fpointnet} use a discrete-continuous loss in the form of $B$ angle bins with regressions within each bin. However, we hypothesize that this divides the training data as each bin will only have a fraction of the total number of training samples to learn from. As shown in Sec.~\ref{sec_ablation_angle_reg}, the network trained in this way does not work as well as when using the angle vector formulation. This is likely due to the small number of pedestrians instances available for training. In our network, the orientation is instead predicted as two values in a vector format, $(x_{\theta}, y_{\theta})= (\cos(\theta),\sin(\theta))$, in the same manner as~\cite{ku_avod}. This formulation avoids dividing the training samples and handles angle wrapping as each angle $\theta$ is represented as a unique unit vector. At inference time, the ROI images corresponding to each object are processed through the network as a single batch to produce $N$ orientation estimates. The final yaw estimation $\theta_{f}$ is calculated as the mean of the angle vector yaw estimates as:

\begin{equation}
\theta_{f} = \tan^{-1}\left(\frac{\frac{1}{N}\sum_{j=1}^{N} y_{\theta_{j}} }{ \frac{1}{N}\sum_{j=1}^{N} x_{\theta_{j}}}\right).
\end{equation}

\subsection{Final Pose Estimation}
The final 3D detection is parameterized by the object centroid $T$, dimensions $D$, and orientation $O$. The centroid and dimension estimates are taken directly from the 3D detections, while the yaw $\theta$ of the orientation is set as the output produced by the Orientation Estimation module.

\subsection{False Positive Suppression}
Pedestrian LiDAR-based 3D detectors' average precision (AP) curves on the KITTI test benchmark show significantly lower precision compared to methods that incorporate image data. This supports the hypothesis that it is significantly harder to discern smaller objects like pedestrians from only LiDAR data, and that false positives are prevalent. To improve detection performance, the 2D boxes from a robust 2D detector are matched with the 2D projections of the 3D detections using an intersection over union (IoU) threshold of 0.4. 3D boxes whose projections do not meet this threshold have their scores reduced, which acts as false positive suppression. Since the KITTI Average Orientation Similarity (AOS) evaluation is also dependent on 2D detection performance, the projection of each 3D box is replaced with its corresponding 2D box from the 2D detector, which allows AOS to be evaluated.

\subsection{Training Losses}
Multi-task training has shown to be effective in improving performance of many neural networks. Following the discrete-continuous angle bin formulation from \cite{mousavian_deep3dbox}, we add an auxiliary output layer that produces $B$ heading bins and $B$ angle regressions. The angle bins are trained with a softmax loss while the bin regression and angle vector outputs are trained with a smooth L1 loss. The total loss is computed as:

\begin{equation}
    L = \alpha L_{ang} + \beta L_{cls} + \gamma L_{reg},
\end{equation}
where $L_{ang}$ is the angle vector loss, $L_{cls}$ is the angle bin classification loss, and $L_{reg}$ is the angle bin regression loss. Losses are weighted such that they converge to values of similar magnitude. We use values of ${\alpha=50}$, ${\beta=1}$, and ${\gamma=200}$. The output from the angle bins are not used in the final estimation as they provide lower performance when used alone. However, incorporating these bins for auxiliary losses in the multi-task training scheme provides additional orientation estimation performance as shown in Sec.~\ref{sec_ablation_angle_reg}.

\section{Implementation Details}

\subsection{Detector Selection}
We use the publicly available AVOD-FPN~\cite{ku_avod} 3D object detector to generate 3D detections. The selection of this detector is based on the recall analysis in Sec.~\ref{sec_recall_analysis}, which shows that it achieves the highest recall for pedestrians compared to other open-source 3D detectors. The open-source 2D detector MS-CNN~\cite{cai_mscnn} is used for false positive suppression. However, our method is compatible with any 3D or 2D detector.

\subsection{View Synthesis}
For each detected object, the camera viewpoints are set to $N=11$ evenly spaced locations ranging from $\rho \in [\ang{-25}, \ang{25}]$ relative to the viewing angle of the object, at a distance {$r=4$ m} as shown in Fig.~\ref{fig_cam_placement}. These minimum and maximum angles are empirically chosen to avoid viewing unseen portions of the pedestrians, which are not captured in the LiDAR scan. The ROI images of size $H \times W$ are generated at a resolution of $224 \times 224$. All 11 images are efficiently rendered in parallel using the Visualization Toolkit (VTK).

\begin{table}[t!]
	\centering
	\begin{tabular}{|l||c|c|c|}
	    \hline
	    \multicolumn{1}{|c||}{Method} & \multicolumn{3}{c|}{Recall} \\
	    \cline{2-4} & Easy & Moderate & Hard \\
	    \hline
        SECOND\cite{yan2018second}   & 86.33 & 80.22 & 73.11 \\
        F-PointNet\cite{qi_fpointnet} & 95.15 & 88.50 & 80.66 \\
        AVOD-FPN\cite{ku_avod}       & \textbf{96.47} & \textbf{93.90} & \textbf{91.16} \\
	    \hline
	\end{tabular}
	\caption{\textbf{3D Pedestrian Detector Recall Analysis.} Recall on moderate pedestrians for top 16 detections at a distance of 0.3~m on the \textit{validation} split. We determine that AVOD-FPN~\cite{ku_avod} has the highest recall of the available open-source 3D detectors.}
	\label{tab_recall_analysis}
\end{table}

\begin{table}[t]
	\centering
	\begin{tabular}{|c|c|c|c|c|}
	    \hline
	    Bins &     Vector &  OS \\
        \hline
           4 &          - & 0.8817 \\
	       8 &          - & \textbf{0.8940} \\
	      12 &          - & 0.8892 \\
	      16 &          - & 0.8625 \\
	    \hline
	       - & \checkmark & 0.9060 \\
	       8 & \checkmark & \textbf{0.9148} \\
	    \hline
	\end{tabular}
	\caption{\textbf{Angle Regression Ablation.} Comparison of Orientation Score (OS) performance for different types of angle regression.}
	\label{tab_ablation_ang_reg}
\end{table}

\begin{table}[th!]
	\centering
	\begin{tabular}{|l||c|c|c|}
	    \hline
	    \multicolumn{1}{|c||}{Viewpoint} & Views & Time (s) & OS   \\
	    \hline
	    ROI Crop    &     1 & 0.042 & 0.8909 \\
	    Virtual Cam &     1 & 0.042 & 0.8692 \\
	    Virtual Cam &     3 & 0.111 & 0.9001 \\
	    Virtual Cam &     6 & 0.209 & 0.9074 \\
	    Virtual Cam &    11 & 0.378 & \textbf{0.9148} \\
	    Virtual Cam &    15 & 0.520 & 0.9061 \\
	    \hline
	\end{tabular}
	\caption{\textbf{Virtual Multi-View Ablation.} Effect of ROI versus virtual camera, the number of multi-views on CNN inference time for 16 objects, and orientation score (OS) when used for training and inference.}
	\label{tab_ablation_multiview}
\end{table}

\subsection{Orientation Estimation Module}
For faster training and convergence, the backbone of the network consists of a modified ResNet-101~\cite{resnet} architecture, taking only the layers before \textit{conv5\_1}, pretrained on the KITTI 2D detection dataset. To reduce GPU memory usage, a 1x1 convolutional layer is applied to the final feature map layer to reduce the depth to 256. The output of this layer is flattened and passed through two fully connected layers of size $(512, 256)$, followed by the orientation estimation output heads. The network is trained on batches of 32 ROI images, while inference can run on up to 16 objects (176 ROI images) simultaneously. The network is trained using an Adam optimizer for 20 epochs with an initial learning rate of 0.0001, which is decayed to 0.00005 at 10 epochs.

\begin{table*}[th!]
	\centering
	\begin{tabular}{|l||c|c|ccc||ccc||ccc|}
		\hline
		\multicolumn{1}{|c||}{Method} & Input & Output & \multicolumn{3}{c||}{2D AP} & \multicolumn{3}{c||}{AOS} & \multicolumn{3}{c|}{OS} \\
		\cline{4-12} & & & Easy & Moderate & Hard & Easy & Moderate & Hard & Easy & Moderate & Hard \\
		\hline
		3DOP\cite{chen_3dop}      & RGB+Stereo & 2D & 81.78 & 67.47 & 64.70 & 72.94 & 59.80 & 57.03 & 0.8919 & 0.8863 & 0.8815\\
		SubCNN\cite{xiang_subcnn} & RGB        & 2D & \textbf{83.28} & \textbf{71.33} & 66.36 & 78.45 & 66.28 & 61.36  & 0.9420 & 0.9291 & 0.9247 \\
		Pose-RCNN\cite{braun2016posercnn}  & RGB+LiDAR & 2D & 77.53 & 63.40 & 57.49 & 73.95 & 59.90 & 54.27 & 0.9538 & 0.9448 & 0.9440 \\
		\hline
        SECOND\cite{yan2018second}& LiDAR      & 3D & 65.73 & 55.74 & 49.08 & 51.56 & 43.51 & 38.78  & 0.7844 & 0.7806 & 0.7901 \\
        AVOD-FPN\cite{ku_avod}    & RGB+LiDAR  & 3D & 67.32 & 58.42 & 57.44 & 53.36 & 44.92 & 43.77  & 0.7926 & 0.7689 & 0.7620 \\
		Ours                      & RGB+LiDAR  & 3D & 81.11 & 70.89 & \textbf{67.23} & \textbf{78.57} & \textbf{67.66} & \textbf{63.83}  & \textbf{0.9687} & \textbf{0.9544} & \textbf{0.9494} \\
		\hline
	\end{tabular}
	\caption{\textbf{Pedestrian Pose Estimation.} \emph{$AP_{2D}$}, \emph{$AOS$}, and Orientation Similarity ($OS$) on the KITTI \emph{test} set. $OS$ is $AOS/ AP_{2D}$. Our pose estimation outperforms all other methods.}
	\label{tab:kitti_test_2d}
\end{table*}

\begin{table*}[t]
	\centering
	\begin{tabular}{|l||ccc||ccc|}
		\hline
		\multicolumn{1}{|c||}{Method} & \multicolumn{3}{c||}{BEV AP} & \multicolumn{3}{c|}{3D AP} \\
		\cline{2-7}                  & Easy & Moderate &  Hard    & Easy & Moderate &  Hard \\ \hline
        SECOND\cite{yan2018second}    &  55.10 & 46.27 & 44.76    & 51.07 & 42.56 & 37.29 \\
        F-PointNet\cite{qi_fpointnet} &  58.09 & 50.22 & 47.20    & 51.21 & 44.89 & 40.23 \\
        AVOD-FPN\cite{ku_avod}        &  58.75 & 51.05 & 47.54    & 50.80 & 42.81 & 40.88 \\
		Ours                          & \textbf{61.46} & \textbf{51.73} & \textbf{47.69} &
		\textbf{53.98} & \textbf{45.01} & \textbf{41.72} \\
		\hline
	\end{tabular}
	\caption{\textbf{3D Pedestrian Localization and Detection.} \emph{$AP_{BEV}$} and \emph{$AP_{3D}$} AP on the KITTI \emph{test} set. Our method outperforms the previous state-of-the-art.}
	\label{tab:kitti_test_3d}
\end{table*}

\section{Experiments and Results}
We validate our method on the challenging KITTI Object Detection benchmark, where we evaluate our results on the Orientation, 3D, and Bird's Eye View (BEV) object detection tasks. We compare with current state-of-the-art methods across all tasks, and show performance improvement across all categories. We follow the common training and validation split used by other methods~\cite{ku_avod, qi_fpointnet, chen_3dop, ku2019monopsr}, which split the training data at a roughly 1:1 training to validation ratio. As with \cite{ku_avod, qi_fpointnet}, a separate training split is used for test submission for better generalization. For training, ROI images are generated from the ground truth centroid annotations, and are augmented with contrast, brightness, and left-right flipping. To simulate 3D localization errors from the 3D detection network, 2D translation is sampled from a uniform distribution up to 10 pixels in any direction. In our ablation studies, we validate our method using renders centered around the top 16 scoring pedestrian centroids per scene generated by AVOD-FPN~\cite{ku_avod}. Experiments are compared using Orientation Score~\cite{mousavian_deep3dbox} (${OS=AOS/AP_{2D}}$), which is the AOS normalized by detection AP. For inference on the test data, we also use the 3D detections from AVOD-FPN. We follow the standard KITTI evaluation difficulty categories of easy, moderate, and hard to evaluate our estimations.

\subsection{Object Recall Analysis}
\label{sec_recall_analysis}
For our method, the selection of the 3D detector is based on the need for high recall as a robust 2D detector can provide strong false positive rejection. Recall analysis for pedestrians based on 3D box IoU can be affected greatly by the initial pose estimation since the bounding boxes are very small. Since we only care about the centroid location and we note that errors along the vertical direction are usually minimal, or less than 8~cm on average, we use an alternative recall metric that evaluates the centroid distance at a threshold of 0.3~m along the horizontal and depth directions, $x$ and $z$, to evaluate whether an object has been detected.

\begin{figure*}[t!]
    \centering
    \begin{tabular}{@{}c@{}}
        \includegraphics[width=0.78\linewidth]{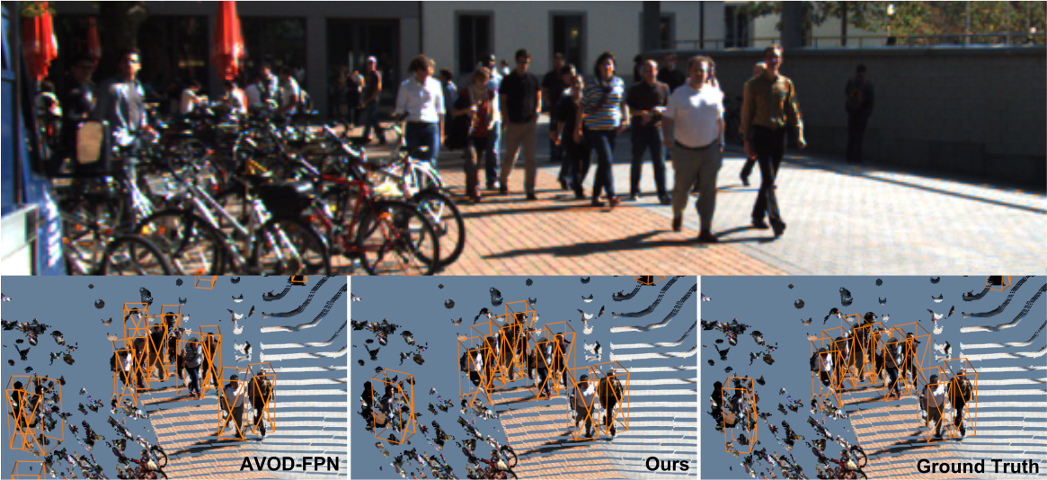}
    \end{tabular}
    \begin{tabular}{@{}c@{}}
        \includegraphics[width=0.78\linewidth]{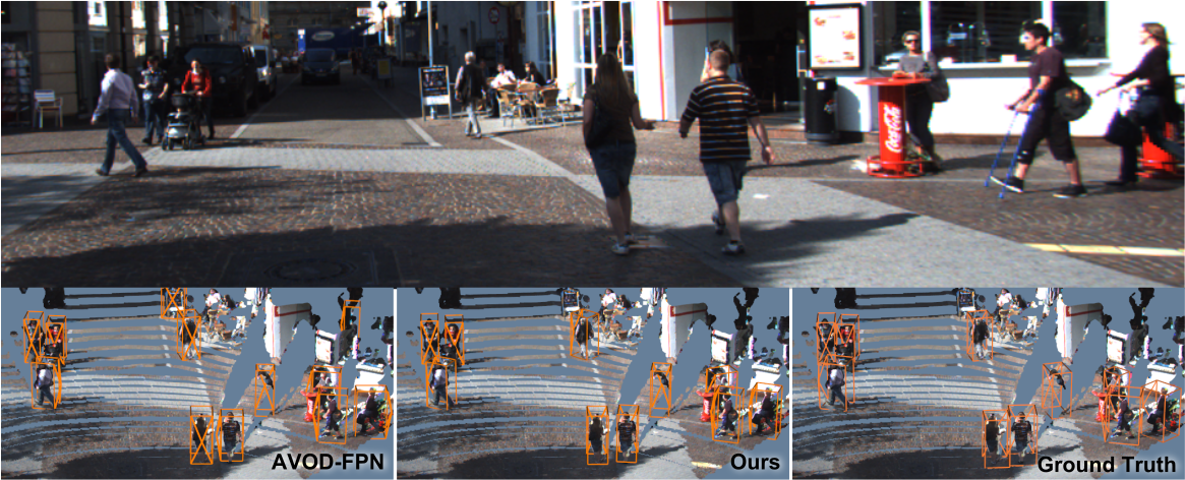}
    \end{tabular}
    \caption{\textbf{Qualitative Results.} We show a comparison of AVOD-FPN detections with and without our orientation module. From left to right: AVOD-FPN~\cite{ku_avod}, Ours, Ground Truth. AVOD-FPN detects all pedestrian instances, but the orientation estimation is poor for several objects and also includes false positives in the detections. Our method estimates orientations that more closely match the ground truth, while also removing false positives.}
    \label{fig_qual_results}
\end{figure*}

In Tab.~\ref{tab_recall_analysis}, we evaluate the recall of three pre-trained open-source 3D detection methods using the top 16 scoring pedestrian detections per scene on the validation split. We find that the LiDAR only method, SECOND, misses a larger portion of instances, with a recall of 80.22\% on the moderate pedestrian instances, while the fusion based AVOD-FPN detector provides the highest recall of 93.90\% on moderate pedestrians. F-PointNet~\cite{qi_fpointnet} also provides a similarly high recall of 88.50\% on moderate, but the 2D detector used in their method is not publicly available.

\subsection{Angle Regression} \label{sec_ablation_angle_reg}
Tab.~\ref{tab_ablation_ang_reg} investigates the effect of using each type of orientation representation. The table shows that the angle bin estimation provides better results at 8 bins compared to 12 or 16, which aligns with the intuition that splitting the training data for each bin results in worse performance. However, using only 4 bins requires the network to learn larger regression values, which does not work as well. The angle vector formulation outperforms the angle bin method, and the final row of the table shows that using the multi-task loss formulation with the angle bins as an auxiliary loss provides the best result.

\subsection{Effect of Rendered Views}
Tab.~\ref{tab_ablation_multiview} shows the influence of the rendered views on the CNN inference time and orientation estimation performance. The basic ROI crop provides better results than a single virtual camera, likely due to the fact that the novel views are being rendered from sparser data. However, with only 3 views, the multi-view formulation outperforms the ROI crop. Performance increases with the number of rendered views, but the improvement comes at the cost of increased inference time on the GPU. However, at 15 views performance no longer improves likely due to the fact the novel views are rendered too close together resulting in repeated data. The final version of the pipeline uses 11 views, which provides the best OS performance. The rendering of 11 views only takes an additional 5~ms over a single view, but each additional view takes 34~ms to inference in the network.

\subsection{Comparison with State-of-the-Art}
We apply our method on the unseen data of the test samples, and submit the results to the online server for evaluation. In Tab.~\ref{tab:kitti_test_2d}, we compare our method with existing methods on the tasks of 2D detection and orientation estimation. On the easy and moderate difficulties, our proposed method achieves comparable 2D AP performance to other 2D methods that also provide orientation estimation, while performing better on hard instances due to the high recall of our selected 3D detector. On the task of orientation estimation, evaluated as AOS on the test server, our method outperforms all other methods on the benchmark across all difficulties, and produces the highest OS. We also compare our results to other 3D detection methods that provide orientation estimates and show significant improvements in orientation estimation performance.

In Tab.~\ref{tab:kitti_test_3d}, we compare our results with the existing state-of-the-art 3D detections methods on the tasks of 3D localization $AP_{BEV}$ and 3D detection $AP_{3D}$. As expected, our performance improves upon the base AVOD-FPN detector due to false positive rejection from MS-CNN, as well as the more accurate orientation estimation which increases 3D and BEV IoU due to better 3D box alignment.

\subsection{Timing}
We assess the run time of each component in our proposed method. For depth completion, we use the multi-scale version of \cite{ku_ipbasic} which takes 33~ms per LiDAR scan. The View Synthesis module takes 40~ms to render 11 views for 16 objects with VTK. On the KITTI dataset, the CNN inference takes 72~ms per frame on average on a Titan Xp GPU using the AVOD-FPN 3D detections, which equates to 24~ms per object. The AVOD-FPN 3D detector runs at 100~ms per frame. In total, the average time per frame is 245~ms. However, it is important to note that our pipeline does not affect the initial object detection time, which depends on the 3D object detector used. Even so, the full estimation is still 8x faster than other methods with comparable pose estimation performance; SubCNN~\cite{xiang_subcnn} and Pose-RCNN~\cite{braun2016posercnn} take 2~s, and 3DOP~\cite{chen_3dop} takes 3~s per frame.

\section{Qualitative Results}
Fig.~\ref{fig_qual_results} shows a qualitative comparison of pose estimation results on the KITTI dataset. AVOD-FPN detects all pedestrian instances within the scene. However, orientations are poorly estimated for several instances and there are also false positives. Our module successfully recovers more accurate orientation estimations for the pedestrians within the scene, and suppresses several false positives.

\section{Conclusion}
This work presents a method to generate virtual viewpoints on a depth completed point cloud fused with RGB data that addresses the issues of learning semantic information from sparse LiDAR data, and learning from small amounts of pose training data. The generation of virtual views in canonical viewpoints also preserves object appearance. The accurate estimation of pedestrian pose is especially important for safe autonomous driving as the behavior of pedestrians can be more easily predicted. Our module is designed to be incorporated into existing 3D detectors to provide accurate pose estimation, and sets new state-of-the-art results on the KITTI object detection benchmark.

\addtolength{\textheight}{-2cm}








{
\bibliographystyle{IEEEtran}
\bibliography{root}
}

\end{document}